# The Evolving Landscape of Generative Large Language Models and Traditional Natural Language Processing in Medicine

Rui Yang[1,2†], Huitao Li[1,2†], Matthew Yu Heng Wong[3], Yuhe Ke[4], Xin Li[1,2], Kunyu Yu[1,2], Jingchi Liao[1,2], Jonathan Chong Kai Liew[1,2,5], Sabarinath Vinod Nair[1,2], Jasmine Chiat Ling Ong[2,6], Irene Li[7], Douglas Teodoro[8], Chuan Hong[9,10], Daniel Shu Wei Ting[11,12,13], Nan Liu[1,2,14,15*]

*†: co-first authors   *: corresponding author*

[1] *Center for Quantitative Medicine, Duke–NUS Medical School, Singapore, Singapore*

[2] *Duke-NUS AI + Medical Sciences Initiative, Duke-NUS Medical School, Singapore, Singapore*

[3] *School of Clinical Medicine, Fitzwilliam College, University of Cambridge, England, UK*

[4] *Division of Anesthesiology and Perioperative Medicine, Singapore General Hospital, Singapore, Singapore*

[5] *School of Chemistry, Chemical Engineering and Biotechnology, Nanyang Technological University, Singapore, Singapore*

[6] *Division of Pharmacy, Singapore General Hospital, Singapore, Singapore*

[7] *Graduate School of Engineering, The University of Tokyo, Tokyo, Japan*

[8] *Department of Radiology and Medical Informatics, Faculty of Medicine, University of Geneva, Geneva, Switzerland*

[9] *Department of Biostatistics and Bioinformatics, Duke School of Medicine, Durham, NC, USA*

[10] *Duke Clinical Research Institute, Durham, NC, USA*

[11] *Singapore Eye Research Institute, Singapore National Eye Center, Singapore, Singapore*

[12] *Byers Eye Institute, Stanford University, Stanford, CA, USA*

[13] *Artificial Intelligence Office, Singapore Health Services, Singapore, Singapore*

[14] *Programme in Health Services and Systems Research, Duke–NUS Medical School, Singapore, Singapore*

[15] *NUS Artificial Intelligence Institute, National University of Singapore, Singapore, Singapore*

**Abstract**

Natural language processing (NLP) has been traditionally applied to medicine, and generative large language models (LLMs) have become prominent recently. However, the differences between them across different medical tasks remain underexplored. We analyzed 19,123 studies, finding that generative LLMs demonstrate advantages in open-ended tasks, while traditional NLP dominates in information extraction and analysis tasks. As these technologies advance, ethical use of them is essential to ensure their potential in medical applications.

**Main Text**

In medicine, large volumes of unstructured textual data are generated daily, such as electronic health records and biomedical literature[1]. Efficiently extracting and structuring this information is crucial for improving healthcare quality, supporting clinical decision-making, and advancing medical research[1]. To achieve this goal, natural language processing (NLP) techniques have been widely applied to medical text mining tasks, including event extraction, trial matching, entity normalization, and diagnostic coding[2,3]. With ongoing technological advancements, NLP has progressed from rule-based systems to statistical learning, and more recently to deep learning approaches, significantly advancing the structuring of medical information and knowledge discovery[2–4].

The rise of large language models (LLMs) is reshaping the development of NLP, demonstrating powerful capabilities in text understanding, generation, and reasoning[5–8]. Unlike traditional NLP methods[1], which rely on large amounts of annotated data and are tailored for specific tasks, LLMs leverage large-scale pre-training data to achieve strong generalization across various tasks[5]. In medicine, LLMs have shown extensive potential in clinical consultation, disease diagnosis, health management, medical education, and medical research[9,10].

The emergence of generative LLMs has introduced novel technological pathways for medical NLP, expanding the scope of applications across various tasks. However, it remains unclear whether their research focus differs from traditional NLP methods. To address this gap, we conducted topic modeling on relevant studies to compare task distributions and application contexts across technological paradigms, aiming to map the current research landscape and inform future directions. Figure 1 illustrates the overall workflow of this study.

---

[1] In this study, we define traditional NLP methods as a continuously period spanning from the 1990s to the 2010s, including early rule-based and statistical approaches (such as bag-of-words, TF-IDF) and more sophisticated models (Hidden Markov Models and Conditional Random Fields), to neural language models that emerged in the early 2010s (such as word2vec, Recurrent Neural Networks, Long Short-Term Memory networks), and to encoder-based models built on the Transformer architecture (BERT series) that appeared after 2017.

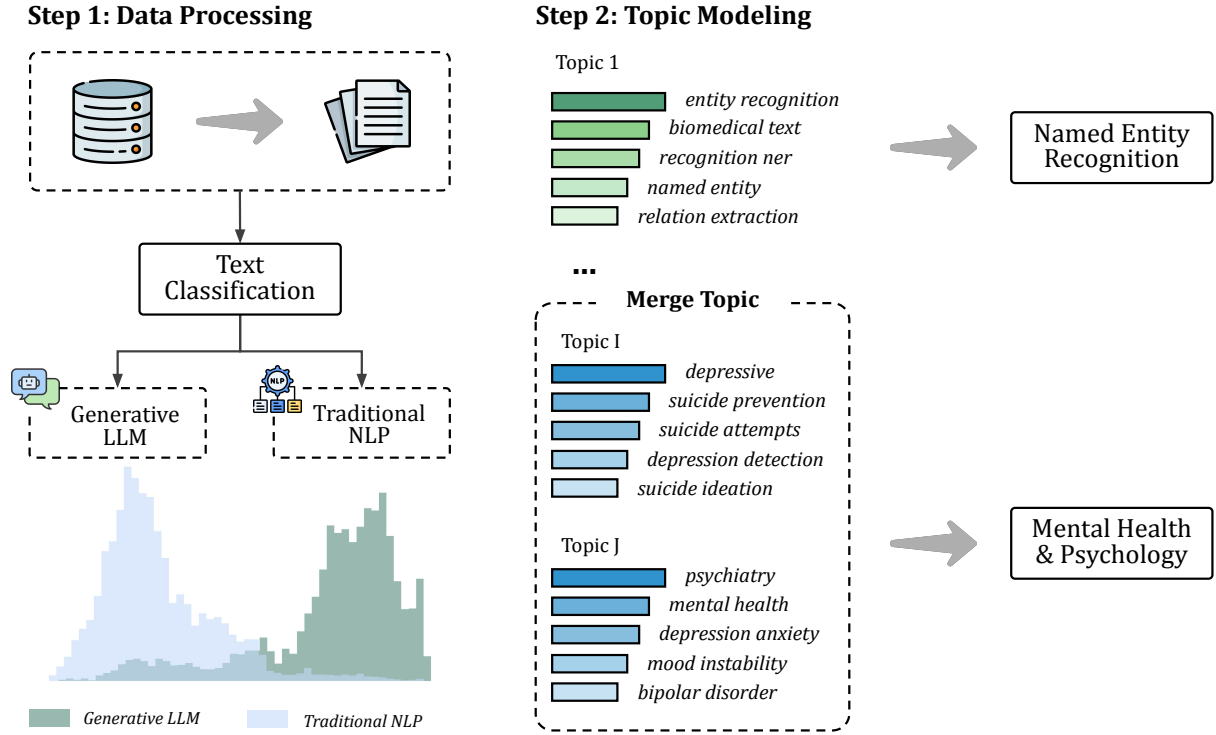

**Figure 1. Study workflow overview.** We retrieved relevant studies from PubMed, Embase, Scopus, and Web of Science, and categorized them into generative LLM and traditional NLP groups based on keywords. We then performed topic modeling on all articles and merged similar topics.

We identified a total of 44,609 studies from PubMed, Embase, Scopus, and Web of Science. After removing duplicates, 20,228 unique studies remained. We excluded 1,105 records that did not meet the criteria and ultimately analyzed 19,123 studies. Among them, 4,295 were categorized as studies related to generative LLMs, while the remaining 14,828 were categorized as studies related to traditional NLP. The PRISMA flow diagram is provided in Supplementary Information A.

To explore the distribution of generative LLM and traditional NLP studies within the semantic space, we embedded all studies using the "MedCPT Article Encoder" model[11] and reduced the embeddings to four dimensions using the UMAP (Uniform Manifold Approximation and Projection) algorithm[12]. As shown in Figure 2, the density distributions for each dimension demonstrate distinct disparities between the two groups, with near-complete separation observed in Dimension 1. Generative LLM studies exhibit a more concentrated distribution, whereas traditional NLP studies display a more dispersed pattern.

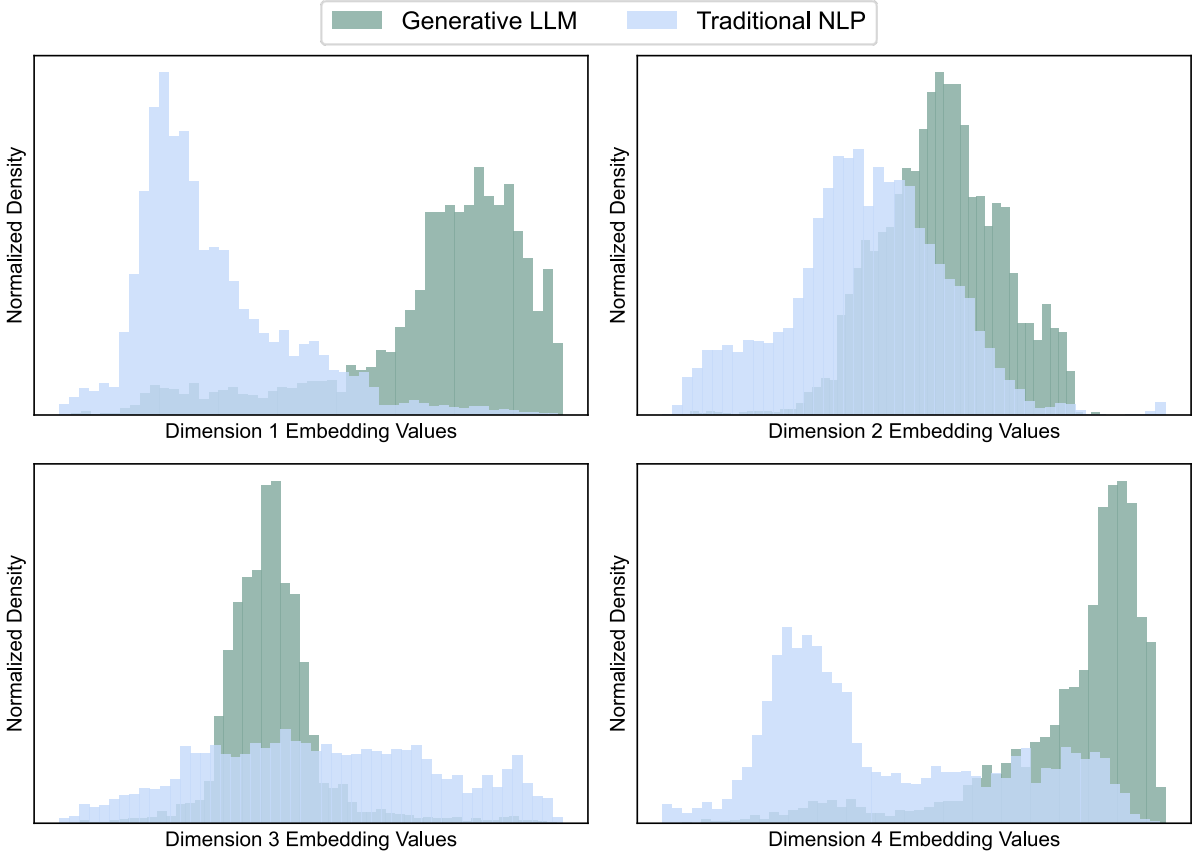

**Figure 2. Semantic embeddings for generative LLM and traditional NLP studies.** All studies were embedded using the "MedCPT Article Encoder" model and reduced to four dimensions via the UMAP algorithm.

Subsequently, we utilized BERTopic[13] to automatically generate 40 initial topics, based on which medical experts merged similar topics, ultimately producing 26 topics. The keywords for all topics are provided in Supplementary Information B. Figure 2 shows the distribution ratio of research related to generative LLMs versus traditional NLP methods across these topics.

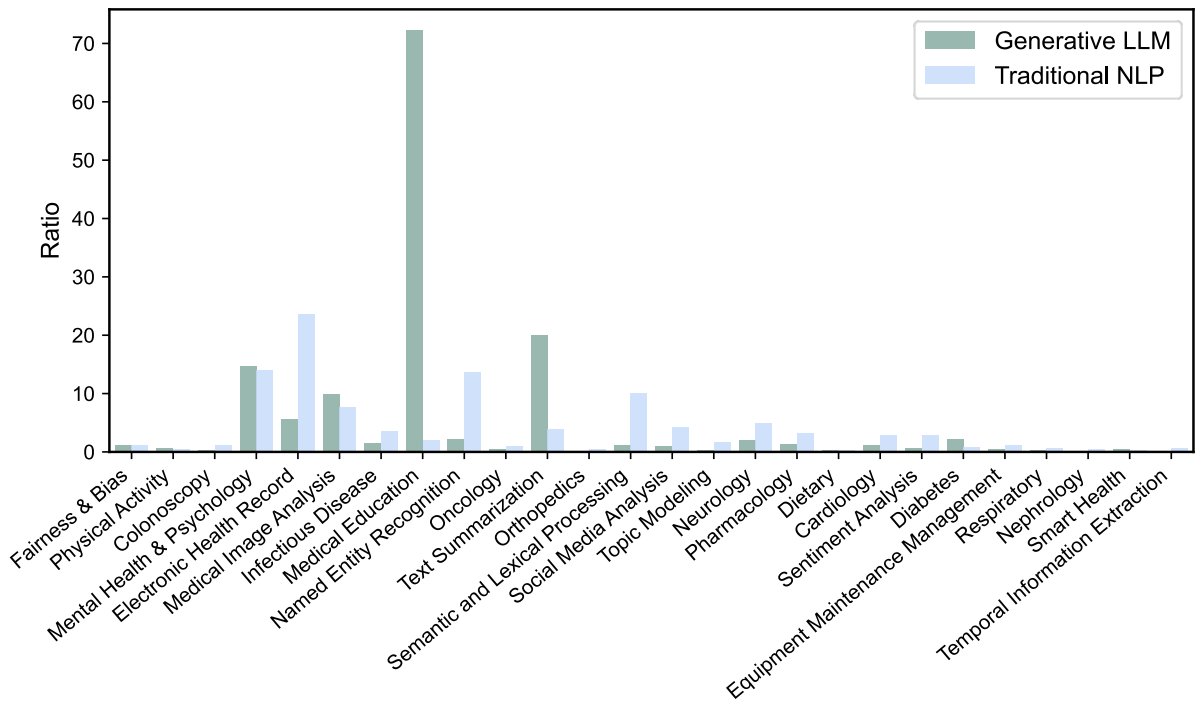

**Figure 3. Distribution of topics across generative LLM and traditional NLP studies.** Generative LLMs exhibit higher concentrations in open-ended tasks such as "Medical Education" and "Text Summarization", while traditional NLP remains dominant in tasks such as "Electronic Health Records" and "Named Entity Recognition".

We observed that generative LLM-related studies showed higher proportions in topics such as "Medical Education" (72.23%), "Text Summarization" (19.95%), and "Medical Image Analysis" (9.80%), reflecting their potential applications in open-ended content generation and cross-modal analysis tasks. The significant proportion in "Medical Education" suggests that LLMs can provide scalable approaches for knowledge delivery that support more flexible medical training. Such applications include simulating clinical scenarios and guideline-based self-assessment, potentially enhancing the interactivity and personalization of medical education and bringing the learning experience closer to real clinical contexts[14]. The prominence of "Text Summarization" indicates the potential of LLMs to improve clinical efficiency by converting lengthy clinical notes, discharge summaries, and radiology reports into more concise formats. This application may reduce the documentation burden on clinicians and enhance the efficiency of information retrieval[15]. In addition, the relatively high representation of "Medical Image Analysis" reflects the growing trend of multimodal processing in medicine, providing new technological approaches for diagnostic interpretation and reporting, and facilitating the integration of visual and textual information[16].

In contrast, among all traditional NLP studies, 23.62% focused on "Electronic Health Records", 13.70% on "Named Entity Recognition", and 9.95% on "Semantic and Lexical Processing". This aligns with traditional NLP's longstanding focus on structured information extraction and terminology normalization, but also highlights its continued focus on semantic analysis and concept recognition. These tasks typically require models with high controllability and precision—areas where conventional approaches still maintain an advantage[17].

This study reveals the differences in research focus between generative LLMs and traditional NLP methods in medicine, demonstrating their complementary advantages across tasks and application scenarios. As LLMs with stronger reasoning capabilities (such as Gemini 2.5[18], OpenAI-o3[19], and DeepSeek-R1[8]) continue to develop, their applications in clinical reasoning and decision support will receive more attention.

However, the integration of LLMs into real clinical settings remains at an early stage. Although LLMs demonstrate certain medical capabilities, they encounter issues such as incomplete reasoning chains and insufficient explainability when facing complex tasks, limiting their depth of application in clinical practice[20]. At the same time, their continuously enhancing capabilities have raised ethical concerns, including privacy protection and bias control[21]. Moving forward, emphasizing responsible development and deployment of LLMs becomes important, aiming to establish controllable and transparent medical AI systems that help ensure fairness in clinical practice[21].

As these technologies advance, the role of healthcare professionals will also adapt accordingly. LLMs can serve as auxiliary tools to enhance clinical judgment, but the expertise of medical professionals remains indispensable. Therefore, cultivating healthcare professionals' ability to evaluate and effectively use these technologies will become increasingly important. Through systematic research and interdisciplinary collaboration, medical NLP technology has the potential to evolve from information processing tools into more intelligent support systems, bringing new possibilities for medicine. This development process needs to be grounded in patient welfare and clinical practice principles, ensuring that technological progress truly serves to improve medical quality and enhance patient experience.

**Methods**

*Search Strategy*

We conducted a systematic literature search using PubMed, Embase, Scopus, and Web of Science to identify relevant studies published prior to March 1, 2025. The search strategy we used included terms related to "natural language processing" and "large language model". We limited our search to "Title/Abstract" and included only "Article" in "English" focused on "Human". The detailed search strategy can be found in Supplementary Information C.

*Data Processing*

After retrieving all studies from the databases, we implemented a data processing workflow to ensure data quality. First, we performed initial deduplication using EndNote, followed by a secondary deduplication with Rayyan to eliminate all redundant studies. Next, we utilized Python to identify and remove studies without available abstracts. Although we had restricted the document type to "Article" during the search stage, we conducted an additional check through Python to remove any misclassified entries. In addition, we conducted an iterative topic modeling analysis on the research corpus and removed irrelevant articles by detecting anomalous topics.

After data cleaning, we classified all studies into generative LLM-related and traditional NLP-related categories. The classification was performed by matching keywords in the study titles and abstracts to ensure accuracy. The specific keyword list can be found in Supplementary Information D.

*Topic Modeling*

Topic modeling is an unsupervised or semi-supervised learning technique used to automatically extract latent topics from documents. In this study, we utilized the "BERTopic" package to perform topic modeling on the retrieved studies[13]. Specifically, we employed the "MedCPT Article Encoder" model, which was trained on 255 million query-article pairs from PubMed search logs, to generate embeddings[11]. Subsequently, we applied the UMAP algorithm to reduce the dimensionality of the embeddings, preserving semantic information while significantly reducing computational complexity[12]. Next, we utilized the Hierarchical Density-Based Spatial Clustering of Applications with Noise algorithm to cluster the reduced embeddings, automatically determining the number of topics and identifying clusters with similar semantics[12,22]. Finally, we generated topic representations and optimized them using GPT-4o[23]. Medical experts further merged similar topics through keywords.

**Data Availability**

The data used in this study were obtained from PubMed, Embase, Scopus, and Web of Science, which are subscription-based databases. Access to these databases requires institutional or individual subscriptions. Due to licensing restrictions, the full data cannot be publicly shared.

**Code Availability**

The data used in this study is available upon request.

**Acknowledgments**

This work was supported by the Duke-NUS Signature Research Programme funded by the Ministry of Health, Singapore. Any opinions, findings and conclusions or recommendations expressed in this material are those of the author(s) and do not reflect the views of the Ministry of Health.

**Author Contributions**

N.L. and R.Y. conceived the study. R.Y., H.L., X.L., K.Y., J.L., J.C.K.L., and S.V.N. collected data. R.Y. and H.L. conducted data analyses. R.Y., H.L., M.W., and Y.K. drafted the manuscript, with further improvement by I.L., D.T., C.H., D.S.W.T, and N.L. N.L. supervised the study. All authors contributed to the revision of the manuscript and approval of the final version.

**Competing Interests**

D.S.W.T. serves as an Associate Editor and N.L. as an Editorial Board Member for npj Digital Medicine. They played no role in the peer review of this manuscript. The remaining authors declare that there are no other financial or non-financial competing interests.

**Supplementary Information A: PRISMA flow diagram**

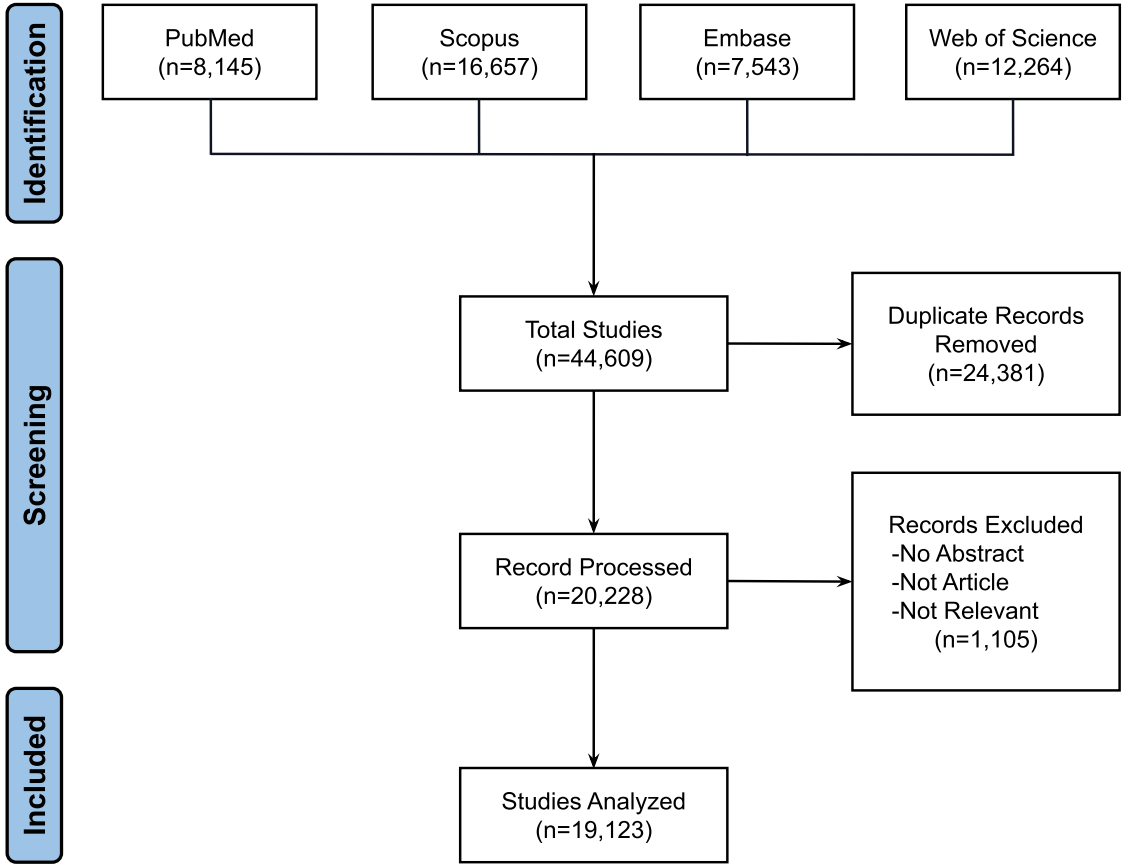

**PRISMA flow diagram for identifying related studies.** Our search retrieved 44,609 study records (n=8,145, 18.26% from PubMed; n=16,657, 37.34% from Scopus; n=7,543, 16.91% from Embase; n=12,264, 27.49% from Web of Science); of these, 20,228 (45.35%) were processed after deduplication (n=24,381, 54.65%). After excluding studies that lacked abstracts, were not articles, or were irrelevant (n=1,105, 5.46%), 19,123 (94.54%) records remained for analysis.

**Supplementary Information B: Topic Keywords**

| ID | Keywords | Topic |
|---|---|---|
| 1 | natural language processing, electronic health records, natural language, health records, electronic health, clinical notes, language processing, clinical, pathology reports, health | Electronic Health Record |
| 2 | examination, readability, exam, patient education, clinical, assessed, medical, patients, medical education, chatbots | Medical Education |
| 3 | entity recognition ner, entity recognition, named entity recognition, biomedical text, relation extraction, recognition ner, named entity, corpus, entities, entity | Named Entity Recognition |
| 4 | sense disambiguation, semantic similarity, disambiguation, information retrieval, semantic, question answering, natural language, lexicon, corpus, lexical | Semantic and Lexical Processing |
| 5 | cnn, medical images, deep learning, medical image, cnns, convolutional neural, medical image segmentation, medical imaging, visual question answering, vision language | Medical Image Analysis |
| 6 | large language models, text summarization, generated summaries, summarizer, language models, extractive summarization, summarization, language models llms, discharge summaries, question answering | Text Summarization |
| 7 | hospitalizations, heart failure, venous thromboembolism, hospitalization, natural language processing, pulmonary embolism, electronic health, natural language, embolism, clinical | Cardiology |
| 8 | drug discovery, antimicrobial, antibacterial, peptides, proteins, peptide, protein language models, protein language model, drug development, immune | Pharmacology |
| 9 | radiology reports, radiology report, radiology, diagnostic accuracy, radiologist, radiologists, radiological, imaging findings, diagnostic, diagnosis | Medical Image Analysis |
| 10 | opioid, veterans, NLP, electronic health records, substance use, PTSD, SBDH, clinical notes, health care, risk factors | Mental Health & Psychology |
| 11 | depression detection, natural language processing, suicide ideation, depressive, deep learning, suicide prevention, social media data, social media posts, depression, suicide attempts | Mental Health & Psychology |
| 12 | vaccine related tweets, vaccine acceptance, covid 19 vaccination, covid 19 vaccine, covid 19 vaccines, vaccine, vaccination, vaccines, vaccine hesitancy, vaccine development | Infectious Disease |
| 13 | mood instability, mental health, depression anxiety, psychiatric, depressive, bipolar disorder, bipolar, depression, psychiatry, psychosis | Mental Health & Psychology |
| 14 | sentiment classification, sentiment scores, sentiment analysis, opinion mining, natural language processing, natural language, based sentiment analysis, user reviews, sentiment, patient reviews | Sentiment Analysis |

| 15 | social media data, tweets, use social media, twitter, social media, tweet, use social, sentiment analysis, sentiment, drug | Social Media Analysis |
|---|---|---|
| 16 | alzheimer dementia, mild cognitive impairment, dementia diagnosis, dementia, diagnosis dementia, alzheimer disease, alzheimer, dementiabank, alzheimer disease ad, cognitive impairment | Neurology |
| 17 | human written abstracts, written abstracts, medical writing, generated abstracts, abstracts generated, human reviewers, scientific writing, medical research, articles generated, original abstracts | Text Summarization |
| 18 | chatbots, conversational agents, conversational agent, social robot, social robots, robot, cognitive behavioral therapy, behavioral therapy, mental health support, human robot | Mental Health & Psychology |
| 19 | electronic health record, electronic health records, electronic health, clinical data, health record ehr, health record, health records, clinical documentation, ehr data, ehr systems | Electronic Health Record |
| 20 | diabetes care, patients diabetes, diabetes mellitus, patients type diabetes, diabetes, type diabetes, type diabetes mellitus, insulin therapy, glycemic control, diet plans | Diabetes |
| 21 | mental health, conversational agents, chatbots, social robots, therapy, human-robot interaction, cognitive behavioral therapy, empathy, support, elderly | Mental Health & Psychology |
| 22 | colonoscopy reports, colonoscopy pathology reports, screening colonoscopy, surveillance colonoscopy, screening colonoscopies, colonoscopy, colonoscopy quality, colonoscopy pathology, colonoscopy examinations, colonoscopies performed | Colonoscopy |
| 23 | primary progressive aphasia, progressive aphasia, fluent aphasia, brain computer interface, aphasia, people aphasia, impairments, brain computer, fmri, aphasic | Neurology |
| 24 | eeg data, eeg dataset, eeg signals, electroencephalogram, electroencephalogram eeg, eeg based, sleep stage classification, electroencephalography eeg, raw eeg, electroencephalography | Neurology |
| 25 | topic modelling, topic modeling, topic models, topic model, lda topic modeling, topic modeling clustering, latent dirichlet allocation, topics, research topics, bibliometric | Topic Modeling |
| 26 | chemotherapy, radiotherapy, radiation therapy, prostate cancer, chemoradiation, cancer, overall survival os, prognostic, metastasis, tumor | Oncology |
| 27 | stigmatizing language, applicant gender, female applicants, male applicants, black patients, white patients, applicants, racial gender, gender bias, racial | Fairness & Bias |
| 28 | child speech, autism spectrum, children autism, autistic children, autism diagnostic, autism spectrum disorder, autism, deaf children, utterances, language development | Neurology |
| 29 | maintenance management, learning models, maintenance activities, hazard recognition, deep learning, neural network, maintenance, forecasting, construction accidents, structural health monitoring | Equipment Maintenance Management |
| 30 | hate speech detection, fake news detection, fake news dataset, abusive language, offensive language, hate speech, cyberbullying, news detection, language hate, bullying | Social Media Analysis |
| 31 | patients epilepsy, epilepsy seizure, patients seizure, seizure outcomes, epileptic seizures, epilepsy, diagnosis epilepsy, epilepsy surgery evaluation, functional seizures, seizures | Neurology |
| 32 | temporal information clinical, temporal information extraction, temporal expression extraction, clinical events temporal, extracting temporal information, temporal relation extraction, extracting temporal, clinical temporal, temporal relations clinical, events temporal expressions | Temporal Information Extraction |
| 33 | covid 19 social, anxiety provoking topics, 19 social distancing, social prescribing, social support, 19 pandemic, mental health, lgbtq teens, covid 19 pandemic, mental health scales | Mental Health & Psychology |
| 34 | medical image retrieval, annotation medical images, images retrieval, image retrieval, image retrieval cbir, based image retrieval, image search, medical images, based retrieval, information retrieval | Medical Image Analysis |
| 35 | asthma criteria, predetermined asthma criteria, asthma ascertainment, asthma care, patients asthma, asthma status, asthma patients, asthma diagnosis, predetermined asthma, diagnosis asthma | Respiratory |
| 36 | food information extraction, food semantic, food composition data, food information, food entities, food health articles, methods food information, food label, food health, food related | Dietary |
| 37 | kidney disease ckd, chronic kidney disease, kidney disease, kidney failure, proteinuria progression, chronic kidney, elevated proteinuria, proteinuria, nephropathy, ckd patients | Nephrology |
| 38 | osteoporosis, potential osteoporotic, osteoporotic fractures, bone health, osteoporotic fracture, osteoporotic, fracture risk, fractures radiology reports, secondary fracture prevention, fracture prevention | Orthopedics |
| 39 | iot, iot based, things iot, security privacy requirements, privacy requirements, internet things, iot systems, internet things iot, reliability iot, security privacy | Smart Health |
| 40 | physical activity intervention, health physical activity, based physical activity, physical activity mental, physical activity, activity intervention, generated exercise recommendations, physical activity pa, physical activity diet, health behavior change | Physical Activity |

## Supplementary Information C: Search Strategy

**NLP and LLM**
"natural language processing" OR "nlp" OR "language model*" OR "chatgpt" OR "llm" OR "bert" OR "retrieval-augmented generation" OR "retrieval augmented generation"

**Medicine**
"health*" OR "medic*" OR "biomedic*" OR "clinic*" OR "hospital" OR "patient*" OR "physician*" OR "doctor*"

PubMed:
Search in **"Title/Abstract", "Journal Article", "English", "Human"**

Web of Science:
Search in **"Topic"** and limit to **"Article", "English"**

Scopus:
Search in **"TITLE-ABS-KEY"** and limit to **"Article", "English"**

Embase:
Search in **"Title or Abstract"** and limit to **"Article", "English", "Humans"**

**Supplementary Information D: Generative LLM-related Keywords**

| Keyword | Regular Expression Used |
|---|---|
| LLaMA | \b(llama)\b |
| Med-PaLM | \b(med[-\s]?palm)\b |
| PaLM | \b(palm)\b |
| Med-Gemini_Gemini | \b(med[-\s]?gemini\|gemini)\b |
| RAG | \b(rag\|retrieval[-\s]?augmented[-\s]?generation)\b |
| LLM | \b(llm\|large language model\|chat[-\s]?gpt)\b |